\documentclass[final,3p,times,12pt]{elsarticle}
\usepackage[T1]{fontenc}
\usepackage{amsmath}
\usepackage{amsfonts}
\usepackage{amssymb}
\usepackage{graphicx}
\usepackage{algorithm}
\usepackage{algorithmicx}
\usepackage{algpseudocode}
\usepackage{booktabs}
\usepackage{array}
\usepackage{multirow}
\usepackage[colorlinks]{hyperref}
\usepackage{cleveref}
\usepackage{xspace}
\usepackage{xcolor}
\usepackage{makecell}
\usepackage{colortbl}
\usepackage{subfigure}
\usepackage{CJK}
\usepackage{color}
\usepackage{mathrsfs}
\usepackage{url}
\usepackage{textcomp}
\usepackage{bm}

\newcommand{\ie}{\emph{i.e.},\ }

\newcommand{\et}{\emph{et al.}\ }

\newcommand{\drugs}{\textit{drugs}\xspace}
\newcommand{\protein}{\textit{protein}\xspace}
\newcommand{\proteins}{\textit{proteins}\xspace}

\newcommand{\MedHop}{\textsc{MedHop}\xspace}
\newcommand{\MEDLINE}{\textsc{Medline}\xspace}
\newcommand{\DrugBank}{\textsc{DrugBank}\xspace}

\newcommand{\reactome}{\textsc{Reactome}\xspace}
\newcommand{\qangaroo}{\textsc{QAngaroo}\xspace}

\newcommand{\TransE}{\emph{TransE}\xspace}
\newcommand{\TransH}{\emph{TransH}\xspace}

\DeclareMathAlphabet{\mathscr}{OT1}{pzc}{m}{it}

\begin{document}

\begin{frontmatter}

\title{Medical Knowledge Graph QA for Drug-Drug Interaction Prediction based on Multi-hop Machine Reading Comprehension}

\author[qfnu]{Peng Gao}
\author[ecnu]{Feng Gao}
\author[net]{Jian-Cheng Ni}
\author[net]{Yu Wang}
\author[hitsz]{Fei Wang}

\address[qfnu]{School of Cyber Science and Engineering, Qufu Normal University, Qufu, China}
\address[ecnu]{School of Computer Science and Technology, East China Normal University, Shanghai, China}
\address[net]{Network and Information Center, Qufu Normal University, Qufu, China}
\address[hitsz]{School of Electronics and Information Engineering, Harbin Institute of Technology, Shenzhen, China}

\begin{abstract}
Drug-drug interaction prediction is a crucial issue in molecular biology. Traditional methods of observing drug-drug interactions through medical experiments require significant resources and labor. This paper presents a medical knowledge graph question answering model, dubbed \emph{MedKGQA}, that predicts drug-drug interaction by employing machine reading comprehension from closed-domain literature and constructing a knowledge graph of ``drug-protein'' triplets from open-domain documents. The model vectorizes the drug-protein target attributes in the graph using entity embeddings and establishes directed connections between drug and protein entities based on the metabolic interaction pathways of protein targets in the human body. This aligns multiple external knowledge and applies it to learn the graph neural network. Without bells and whistles, the proposed model achieved a 4.5\% improvement in terms of drug-drug interaction prediction accuracy compare to previous state-of-the-art models on the \qangaroo \MedHop dataset. Experimental results demonstrate the efficiency and effectiveness of the model and verify the feasibility of integrating external knowledge in machine reading comprehension tasks.
\end{abstract}


\end{frontmatter}

\section{Introduction}\label{sec:intro}

In recent years, molecular biology has undergone rapid development, with research resources in this field, such as drug analysis literature, structure diagrams, and interactive gene analyses, experiencing substantial growth~\citep{drugbank4,drugbank5,review2-1}.
These advancements, however, mean that researchers now have abundant resources to review, investigate and organize crucial information.
Recent achievements in natural language processing (NLP) can accelerate this process and improve the reliability of further research~\cite{review3-1,review3-2}. In the meantime, linguistics could take advantage of the large, well-curated resources as well~\citep{Cohen2004}, benefiting both fields.

Drug-drug interaction (DDI) is a significant problem, sometimes causing patients irreparable harm \citep{rohani2019drug} due to the concurrent consumption of two or more \drugs~\citep{lazarou1998incidence,review2-2,review2-4}.
As a fundamental research issue in the field of molecular biology, DDI has received a lot of great recent attention.  
Considering the importance of DDI in health, industry, and the economy, and the substantial cost and time of experimental approaches~\citep{hanton2007preclinical,rohani2019drug}, it is an ideal setting for automated solutions.
Welbl \et \citep{welbl-etal-2018-constructing} argued that Information Extraction (IE) applications, such as discovering DDI, must move beyond a scenario where relevant information was coherently and explicitly stated within a single document.
They proposed the \qangaroo \MedHop dataset as a solution~\citep{welbl-etal-2018-constructing}. This dataset searches and records the research paper abstracts from \MEDLINE and combines individual observations, sometimes suggesting previously unobserved DDIs~\citep{doi:10.1086/601720}.

Since multi-hop machine reading comprehension (MRC) aims to find the answer to a given question across multiple documents \citep{10.1007/978-3-030-60450-9_3}, we propose it as a way to address the abundance of literature resources available in molecular biology.
Compared to open-domain MRC, the difficulties of molecular biology closed-domain MRC could be divided into two main topics:
\begin{enumerate}
  \item There is insufficient document comprehension ability in professional medicine. To improve prediction accuracy, models are required to learn the biological or chemical properties in the document when reading and reasoning in the professional literature. Compared with open-domain MRC, closed-domain MRC is a rigorous, detailed, and explainable process that requires interpretation of the transporter activities and reasoning logic of DDI from the perspective of multiple documents MRC.
  \item The limited or superficial description in the document hinders comprehensive learning of the biochemical essence of DDIs. Due to the limitation of document scope during reading, it is not conducive to expressing all the attributes of \drugs or \proteins, which limits DDI prediction.
\end{enumerate}

Therefore, understanding the biological or chemical interactions that cause DDI is important for identifying and preventing them.
Alteration of \textit{in vivo} enzymes and/or transporter activities of a new molecular entity by concomitant \drugs may lead to an altered response safety or efficacy \citep{zhang2009predicting}.
Furthermore, understanding all of the components of the \drugs and \proteins is crucial to understanding potential DDI.
Current DDI prediction between two \drugs makes it hard to verify whether the data leading to these predictions include all the attributes of those \drugs. Not including all the attributes of \drugs or \proteins may cause incomplete reaction chains leading to prediction failures.
Any DDI solution should therefore add as much information as possible on all \drugs.
Leveraging external knowledge bases (KBs) is one way to enable context- and knowledge- aware for MRC and question answering (QA) \citep{beida,DBLP:conf/acl/InkpenZLCW18,DBLP:conf/semeval/WangSZSL18}, which could contribute to DDI predictions.

\begin{figure}[tbp]
     \centering
     \includegraphics[width=\textwidth]{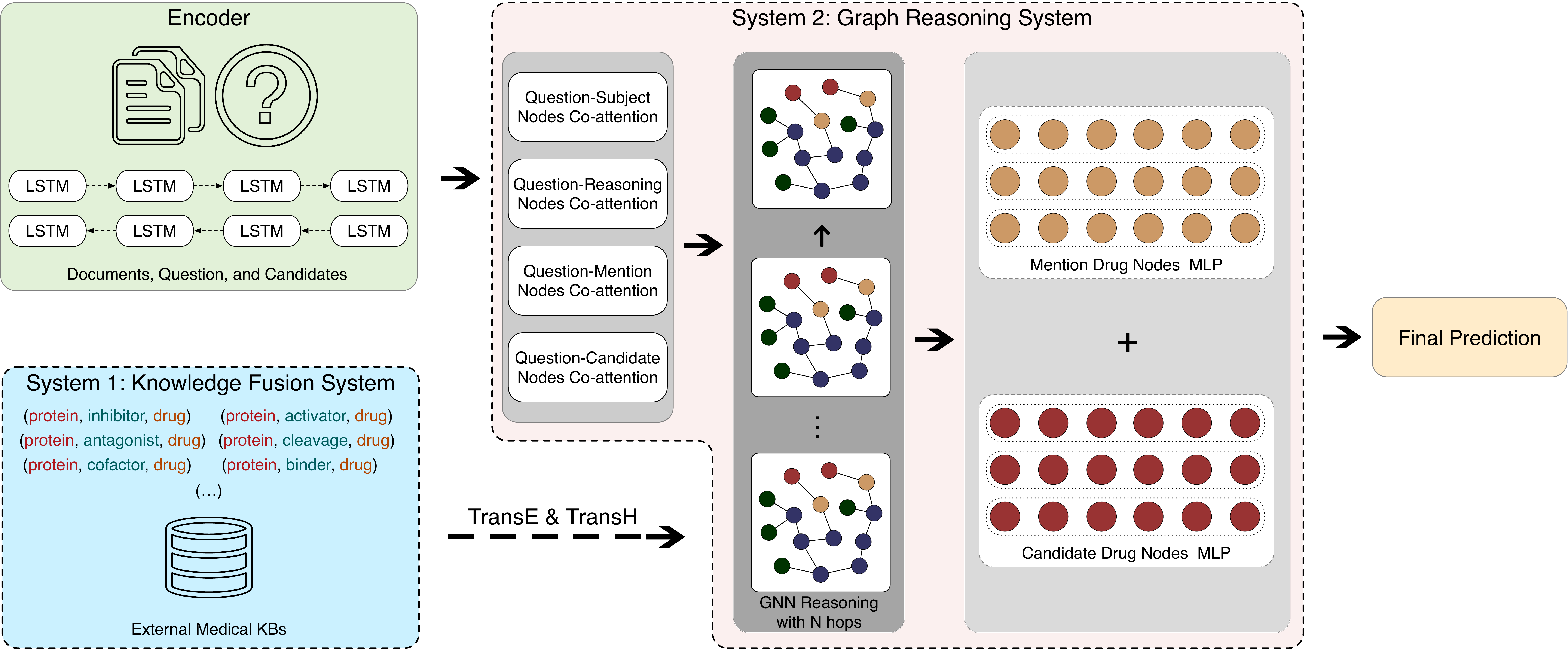}
     \caption{Overview of \emph{MedKGQA}. The arrows indicate the workflows and the color mapping is consistent in this paper: the subject nodes are green, the reasoning nodes are violet, the mention nodes are brown, and the candidate nodes are red. More details about Systems 1 and 2 will described in Sections \ref{sec:sys1} and \ref{sec:sys2}.}
     \label{fig:modelarchitecture}
\end{figure}

In this paper, we propose a model, named \emph{Medical Knowledge Graph Question Answering} (\emph{MedKGQA}), that can address these current challenges. We apply this framework to the \qangaroo \MedHop dataset to predict DDI using an MRC method.
\emph{MedKGQA} is composed of two systems, the knowledge fusion system (System 1) and the graph reasoning system (System 2).
The completed architecture of MedKGQA is illustrated in Figure~\ref{fig:modelarchitecture}.
System 1 is responsible for filling the gap between the potential properties of medical entities and reasoning processing. System 2 is responsible for the reasoning on the directed graphs that contain \drugs and \proteins as nodes and different types of edges following biomedical pathways.

Our contributions can be listed as follows:
\begin{itemize}
    \item We propose the \emph{MedKGQA} model to predict DDI from the medical multi-hop MRC dataset \qangaroo \MedHop, and verify that it achieves state-of-the-art accuracy.
    \item We introduce a methodology that supplements the latent attributes of medical entities through KBs to promote the graph neural network (GNN) reasoning process, which could also be transferred to other closed-domain MRC.
    \item Our model allows us to obtain the visualization and computation results of the reasoning process, which could reflect the interactions among drugs and contribute to further medical research.
\end{itemize}

The rest of the paper is organized as follows.
Section \ref{sec:re} briefly describes the work related to multi-hop MRC and DDI, and Section \ref{sec:model} illustrates the proposed \emph{MedKGQA} framework.
Results and discussions are detailed in Section \ref{sec:expe}.  Section \ref{sec:con} summarizes the paper.

\section{Related Works}\label{sec:re}

\subsection{Multi-document MRC}

Initially, researchers proposed the document selection approach to improve multi-document MRC by leveraging single-document MRC models due to the similarity between these tasks. These approaches involve concatenating multiple documents into an ultra-long concatenated sequence, selecting a single document from multiple documents, and using single-document reading models such as recurrent neural networks (RNNs) \citep{cho2014learning} to perform MRC tasks. Wang \et \citep{wang2018multi} proposed to concatenate the document set into a sequence and applied bi-directional long short-term memory (Bi-LSTM) models \citep{hochreiter1997long,zhou2016attention} to learn squence representations. The authors then extracted a candidate answer from each document subsequence in the concatenated sequence for joint verification, and finally obtained the answer with the highest semantic similarity to the question. To reduce the reading pressure caused by multiple documents on the model, Mao \et \citep{clark2018simple} used pre-trained language models GPT-3 \citep{brown2020language} and BART-large \citep{lewis2020bart} to expand the question statement and improve the accuracy of selecting a single document in the document set. However, concatenating multiple documents using RNN-based models increases training time and the forgetting of previous information. On the other hand, selecting documents from a document set ignores the logical relationship between the documents, reducing the ability for reasoning when reading documents. The attention mechanism is free from the temporal dependence of the text sequence, and its parallel computing advantage reduces the time consumption of learning text representations. Especially after the release of Transformer \citep{transformer}, it has formed a two-stage language model training method of pre-training and fine-tuning. Such models fully utilize the advantages of ultra-high computing power and semantically rich massive literature data, combined with the intrinsic task characteristics of multi-document MRC datasets, to fine-tune the architecture of pre-training language models, and modify the parameter distribution to adapt to new downstream tasks \citep{beltagy2020longformer,zaheer2020big}. Beltagy \et \citep{beltagy2020longformer} proposed the Longformer, a pre-trained language model based on Transformer, which introduces new attention mechanisms to improve the maximum number of tokenized words and document capacity for text sequences. Zaheer \et \citep{zaheer2020big} migrated the graph-oriented sparsification method to the optimization process of Transformer, proposing three mechanisms of random attention, window attention, and global attention, which reduce the computational complexity from exponential to linear, further enhancing the ability to capture semantic features of questions and answers in long texts. However, as the number of documents increases, Transformer-based MRC models still face issues such as upper limits on the number of documents, computational resource bottlenecks, and exploding search space. On the other hand, GNN-based reading models \citep{de-cao-etal-2019-question,cao-etal-2019-bag,cluereader,review2-5} have innovatively extracted spatial topology features through structured methods, providing flexible support for adapting to changing document scales during the reading process.

\subsection{Drug-Drug Interaction Prediction}

DDI prediction research focuses on similarity matrices.
Zhang \et \citep{zhang2015label} proposed an integrative label propagation framework based on the weighted similarity network to predict DDI by integrating clinical side effects, off-label side effects, and chemical structures.
Zhang \et \citep{zhang2017predicting} proposed ensemble models that contain neighbor recommender models, random walk models, and perturbation matrix models, integrating multi-source data that may influence DDI, such as drug substructures, drug targets, drug transporters, drug pathways, drug side effects, and known DDI data, as are flexible frames with suitable ensemble rules for both multi-source features and cooperation between multiple models.
More recently, Rohani \et \citep{rohani2020iscmf} presented an integrated similarity-constrained matrix factorization for DDI prediction using latent features, explicit features for drugs, and known DDIs data as inputs, which can update and learn latent features from the topological structure of graph.
Research on using existing published literature for predicting DDI is limited.
Jiang \et \citep{DBLP:conf/acl/JiangJCB19} proposed a 3-module reading system called Explore-Propose-Assemble reader (EPAr) to select relevant documents and form a tree structure to assimilate information from reasoning chains.
Some solutions \citep{de-cao-etal-2019-question,cao-etal-2019-bag,review2-3} use graph-based open-domain question-answering models that read documents and model the entities into reasoning graphs mainly based on GNNs. The connections of nodes in these models, however, do not include the information indicating how two entities interact.

Inspired by computational methodologies, we propose a dedicated graph-based reasoning model based on medical prior knowledge and evaluate our work in the MRC environment. We hope the proposed model can contribute to the field of NLP as well as medical research.

\section{Method}\label{sec:model}

\subsection{Task Definition}\label{sec:taskdef}

The formalization of the question is denoted as $q=(s, r, a^{*})$, where $s$ is a subject drug, $r$ is the only one relation \textit{interacts\_with}, and $a^{*}$ is the object drug interacting with $s$, and $a^{*}$ will be replaced with an empty slot in prediction.
Since DDI are caused by protein-protein interaction (PPI) chains, we aim to predict the possible correct object drug from the candidate drug set $C$ after reading and reasoning across the independent research paper abstract document set $S$ from \MEDLINE with the help of PPI and present the answer in a multi-hop MRC way.

\subsection{Knowledge Fusion System}\label{sec:sys1}

Translating embeddings for modeling multi-relational data is a powerful method for enhancing link prediction \citep{transe}.
Similarly, we consider DDI reasoning as a link prediction task. With the node pairs, protein-to-protein (\textit{p2p}) and protein-to-drug (\textit{p2d}) and drug-to-drug (\textit{d2d}), are part of DDI prediction chains.

In order to obtain all the biological and chemical natures of drugs and proteins, we collected all currently proven relationships from \DrugBank \citep{drugbank4,drugbank5} and extracted 57 actions between protein and drug, such as \textit{activator, blocker, cofactor}.
Then, \drugs, \proteins, and relationships were recorded as triplets $S$, $(p, l, d)\in S$, the entity set is $E$, the \emph{protein} and \emph{drug} are denoted as $p, d \in E$, and the relation label is $\ell \in L$.

We exploited \TransE \citep{transe} and \TransH \citep{transh}, two effective Knowledge Graph embedding models, to handle the multiple relations and entities in continuous vector space.
We followed the training strategies proposed in Equations~\ref{eq:transe} and \ref{eq:transh}, which both minimize the margin-based ranking criterion \citep{transe,transh}, and reserve the embedded trained medical entities for use in the graph reasoning system.

\begin{equation}\label{eq:transe}
\mathcal{L}_{TransE} = \sum_{(p,\ell,d)\in S} \sum_{(p^{'},\ell,d^{'})\in S^{'}_{(p,\ell,d)}}[\gamma + d(\bm{p} + \bm{\ell}, \bm{d}) - d(\bm{p^{'}}+\bm{\ell},\bm{d^{'}})]_{+}
\end{equation}

\begin{equation}\label{eq:transh}
\mathcal{L}_{TransH} = \sum_{(p,\ell,d)\in S} \sum_{(p^{'},\ell,d^{'})\in S^{'}_{(p,\ell,d)}}[f_{r}(\bm{p},\bm{d}) + \gamma - f_{r^{'}}(\bm{p}^{'},\bm{d}^{'})]_{+}
\end{equation}
where $\gamma$ represents the margin parameter and $d$ represents the similarity measure for $S$, which can be calculated using the Euclidean distance. Equation \ref{eq:transh} calculates the distance based on the hyperplane $f_{r}$. Minimizing Equations~\ref{eq:transe} and \ref{eq:transh} during training updates the vector embeddings of \drugs, \proteins, and relationships. The vector embeddings of \drugs and \proteins obtained from the \TransE and \TransH models are denoted as $\mathbf{U}_{TransE}\in\mathbb{R}^h$ and $\mathbf{U}_{TransE}\in\mathbb{R}^h$, respectively.
Three independent Bi-LSTM \citep{lstm} are next employed to encode documents set $S$, question $q$, and candidate $C$.
Outputs of encoders are $\mathbf{H}_{s}^{i}\in \mathbb{R}^{l_{s}^{i}\times{h}}$, $\mathbf{H}_{q}\in\mathbb{R}^{l_{q}\times h}$, and $\mathbf{H}^{j}_{c}\in \mathbb{R}^{1\times{h}}$ respectively, where $l_{s}^{i}$, $l_{q}$ represent the length of the \textit{i}-th document and question, $j$ stands for the $j$-th drug in $C$, and $h$ is the dimension of Bi-LSTM encoder.

\subsection{Graph Reasoning System}\label{sec:sys2}

\subsubsection{Graph Construction}

\begin{figure}[t]
    \centering
    \includegraphics[scale=0.4]{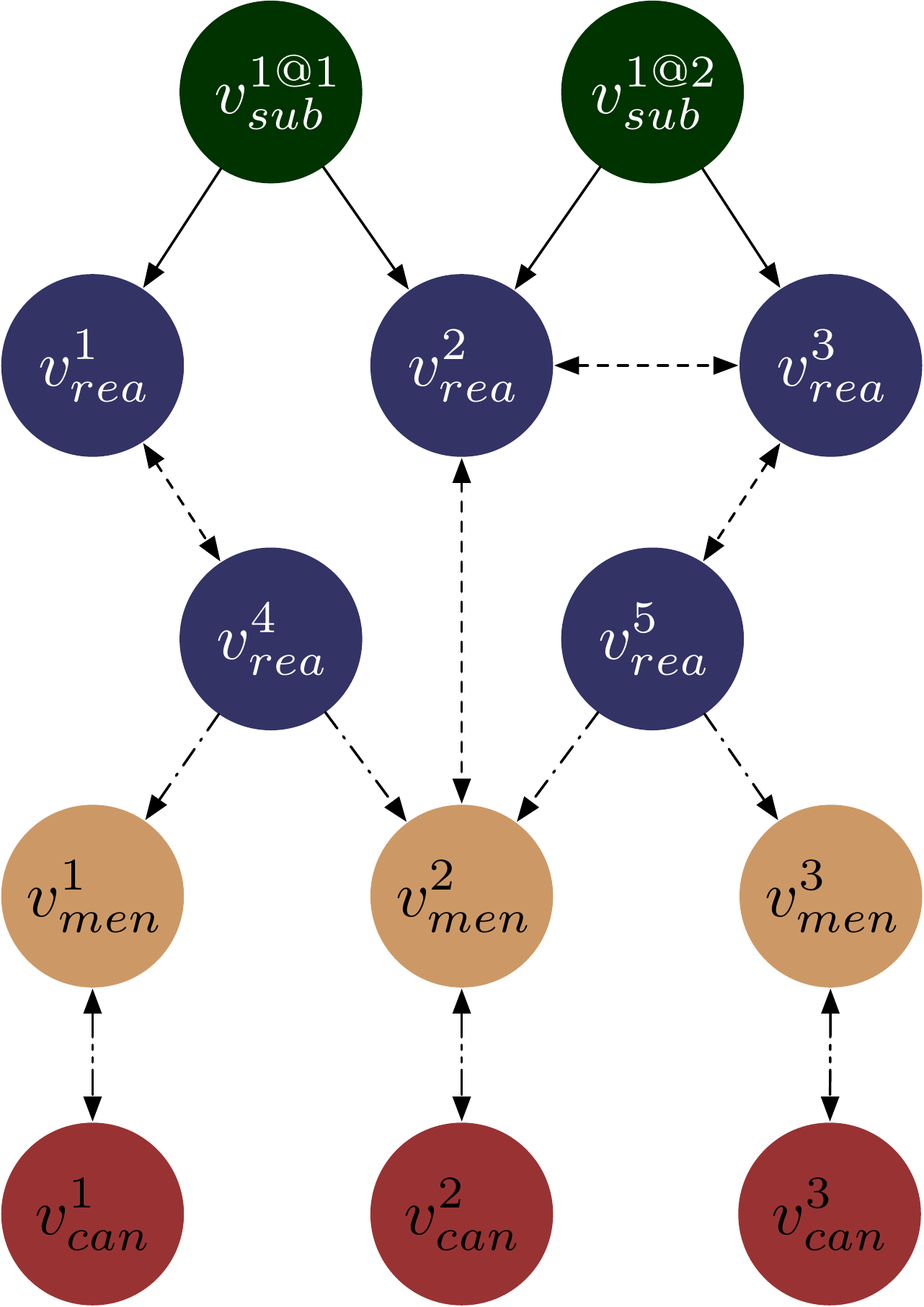}
    \caption{Illustration of the reasoning graph $\mathcal{G}$.
    Subject nodes $\mathcal{V}_{sub}$, mention nodes $\mathcal{V}_{men}$, reasoning nodes $\mathcal{V}_{rea}$, and candidate nodes $\mathcal{V}_{can}$ are colored in green, brown, violet, and red, respectively.
    The symbol @ stands for one entity’s order number appearing in documents (For clarity, only subject nodes are used as examples.)}
    \label{fig:reasonchain}
\end{figure}

The reasoning graph is denoted as $\mathcal{G}=\{\mathcal{V}, \mathcal{E}\}$,  $\mathcal{V}$ stands for the node set on $\mathcal{G}$, and $\mathcal{E}$ contains the edges between nodes.
Figure \ref{fig:reasonchain} illustrates the constructed reasoning graph.

\paragraph{Node Definition} We introduce four kinds of nodes into $\mathcal{G}$: extract text spans matching to subject \drugs $s$ from documents as subject nodes (denoted as $\mathcal{V}_{sub}$ and colored in green in Figure \ref{fig:reasonchain}), text spans matching to \drugs in the candidate set $C$ as mention nodes (denoted as $\mathcal{V}_{men}$ and colored in brown in Figure \ref{fig:reasonchain}), all text spans recognized as \proteins (denoted as $\textbf{P}_{ex}$) from documents as reasoning nodes (denoted as $\mathcal{V}_{rea}$ and colored in violet in Figure \ref{fig:reasonchain}), and \drugs in the candidate set are defined as candidate nodes (denoting as $\mathcal{V}_{can}$ and colored in red in Figure \ref{fig:reasonchain}).

Not all $\textbf{P}_{ex}$ are introduced into $\mathcal{G}$, and they are selected or removed following Algorithm~\ref{alg:1}.
To keep the reasonable DDI and PPI chains, we designed two source-target mapping functions based on prior knowledge: $F(d)$ can confirm whether a $\textit{protein}$ is a target of \textit{drug} $\textit{d}$ and $G(p)$ can confirm whether two $\textit{proteins}$ are covered in PPI chains.

\begin{algorithm}[ht]
    \caption{Iterative Protein Selection Procedure}
    \label{alg:1}
    \begin{algorithmic}[1]
      \Require subject drug $d_{s}$, \textit{d2p} relation $F(d)$, \textit{p2p} relation $G(p)$, extracted protein $\textbf{P}_{ex}$
      \Ensure selected protein $\mathcal{V}_{rea}$
      \State initialize $\textbf{P}_{se}=\emptyset$
      \State get the target \proteins $\textbf{P}_{d_{s}}=F(d_{s})$ of $d_{s}$
      \State $\textbf{P}_{anc}=\textbf{P}_{d_{s}} \cap \textbf{P}_{ex}$
      \Repeat
      \State pop a $p$ from $\textbf{p}_{anc}$
      \If{$p \notin \textbf{P}_{se}$}
      \State add $p$ to $\mathcal{V}_{rea}$
      \State add \proteins interacting with $p$ to $\textbf{P}_{anc}$
      \State $\textbf{P}_{anc}=\textbf{P}_{anc} \cup (G(p) \cap \textbf{P}_{ex})$
      \EndIf
      \Until{$\textbf{P}_{anc}=\emptyset$}
      \State \textbf{return} $\mathcal{V}_{rea}$
    \end{algorithmic}
  \end{algorithm}

\paragraph{Edge Connections} We define four types of edges among nodes, which follow the biochemical reaction pathways in \DrugBank \citep{drugbank5} and \reactome \citep{fabregat2018reactome}:
\begin{enumerate}
    \item $e_{sub2rea}$ --- a directed edge from $v_{sub}$ to $v_{rea}$, if the \protein is known as a target of the \textit{subject drug}.
    \item $e_{rea2rea}$ --- an undirected edge between $v_{rea}$ and $v_{rea}$, if they are covered in a known medical pathway.
    \item $e_{rea2men}$ --- a directed edge from $v_{rea}$ to $v_{men}$, if the \protein is known as a target of the \textit{mention drug node}.
    \item $e_{men2can}$ --- an undirected edge between $v_{men}$ and $v_{can}$, if a \textit{mention node} is a mention of the \textit{candidate drug}.
  \end{enumerate}

\subsubsection{Node and Question Co-attention}
To combine learned question information with nodes extracted from documents, we used co-attention to generate question-aware representations of nodes, which has been shown to be a useful method in previous research~\citep{hde,DBLP:conf/iclr/ZhongXKS19}.
We followed the co-attention implementation, as Tu \et~\citep{hde} outlined.

As an example, we considered the co-attention affinity matrix between the question and subject drug entities, which can be calculated as:

\begin{equation}\label{eq:coattention_q2sub1}
  \mathbf{A}_{qsub}=\mathbf{H}_{sub}(\mathbf{H}_{q})^\top \in \mathbb{R}^{1\times l_{q}}
\end{equation}
where $\top$ stands for matrix transpose.
Each entry of the matrix $\mathbf{A}_{qsub}$ describes how the two words are related between the question and subject nodes $\mathcal{V}_{sub}$.
Then, the question's and $\mathcal{V}_{sub}$'s attention-aware contexts are calculated as:

\begin{equation}\label{eq:coattention_q2sub2}
  \mathbf{C}_q=\mathrm{softmax}((\mathbf{A}_{qsub})^\top)\mathbf{H}_{sub}\in \mathbb{R}^{l_{q}\times h}
\end{equation}

\begin{equation}\label{eq:coattention_q2sub3}
  \mathbf{C}_{sub}=\mathrm{softmax}(\mathbf{A}_{qsub})\mathbf{H}_{q}\in \mathbb{R}^{1\times h}
\end{equation}

The $\mathrm{softmax}(\cdot)$ is used for column-wise normalization, and a Bi-LSTM $f$ is used to encode both $v_{sub}$ and the question co-attended sequence:

\begin{equation}
  \label{eq:coattention_q2sub4}
  \textbf{D}_{sub}=f(\mathrm{softmax}(\mathbf{A}_{qsub})\textbf{C}_q) \in \mathbb{R}^{1\times h}
\end{equation}

Then the node's representation is assigned as the concatenation (denoting as ``$||$'') of $\mathbf{C}_{sub}$, $\mathbf{D}_{sub}$ and knowledge embedding $\mathcal{K}$ from System 1.

\begin{equation}\label{eq:coattention_q2sub5}
  \tilde{\mathcal{U}}_{sub}=[\mathbf{C}_{sub}||\mathbf{D}_{sub}]\in \mathbb{R}^{1\times 2h}
\end{equation}

\begin{equation}\label{eq:coattention_q2sub6}
    \mathcal{U}_{sub}=[\tilde{\mathcal{U}}_{sub}||\mathcal{K}_{\TransE}||\mathcal{K}_{\TransH}]\in \mathbb{R}^{1\times (2h+2d)}
\end{equation}

Equation~\ref{eq:coattention_q2sub1} to \ref{eq:coattention_q2sub6} are also applied to nodes $\mathcal{V}_{men}$, $\mathcal{V}_{rea}$ and $\mathcal{V}_{can}$ to obtain their representations $\mathcal{U}_{men}$, $\mathcal{U}_{rea}$ and $\mathcal{U}_{can}$, respectively.

\subsubsection{Reasoning based on GAT}

We present the reasoning based on Graph Attention Networks (GATs) \citep{gat}, which use node representations and their spatial connections for inter-entity and relation-centric learning.
Empirically, we consider node-pair importance as the indicator to reveal the links' effectiveness in PPI chains and DDI predictions.

\paragraph{Multi-relation Graph Attention Networks} The constructed graph $\mathcal{G}=\{\mathcal{V}, \mathcal{E}\}$ and node representations $\mathcal{U}$ are input to multi-relation GATs.
The attention coefficient between $v_i$ and its neighbor $v_j$ is computed as:

\begin{equation}\label{eq:gat_coattention}
  e_{ij}=a(\mathbf{W} u_i, \mathbf{W} u_j)
\end{equation}
where node representations are initially applied using a weight matrix $\mathbf{W} \in \mathbb{R}^{(2h+2d)'\times (2h+2d)}$, $a: \mathbb{R}^{(2h+2d)'\times (2h+2d)'} \rightarrow \mathbb{R}$ is a shared attentional mechanism, and $e_{ij}$ indicates the \emph{importance} from $v_j$ to $v_i$.
To consider all the \emph{neighbors} of node $i$, the ${\rm softmax}$ function is used for normalizing them across all the choices as:

\begin{equation}\label{eq:gat_softmax}
  \alpha_{ij}={\rm softmax}_j(e_{ij})=\frac{{\rm exp}(e_{ij})}{\sum_{k\in \mathcal{N}_i}{\rm exp}(e_{ik})}
\end{equation}

To adapt to four types of edges, we propose the relational edges modeling as:

\begin{equation}\label{eq:gat_softmax}
    u_i^{l+1}=\frac{1}{K}\|_{k=1}^K \sigma \left( \sum_{j\in \mathcal{N}_{i}} \sum_{r\in \mathcal{R}_{ij}}\frac{1}{\vert\mathcal{N}_{i}^{r}\vert}\alpha_{r_{ij}}^{k,l}\mathbf{W}_{r_{ij}}^{k,l} u_{j}^{l} \right)
\end{equation}
where $ u_{i}^{l}\in \mathbb{R}^{2h+2d}$ is the $i$-th node hidden state in the $l$-th layer, $\mathcal{N}_{i}$ is the neighborhood of $v_i$, $\alpha_{r}^{k,l}$ is the normalized attention coefficients computed by the $k$-th head attention mechanism in relation $r$, and $\|$ indicates the concatenation from $K$ attention mechanisms which is extended from \citep{transformer,gat}.

\paragraph{Gating Mechanisms} To overcome the smoothing problem in node representation updates, we followed the question-aware gating mechanism \citep{DBLP:conf/ijcai/TangSMXYL20} and the general gating mechanism \citep{DBLP:conf/icml/GilmerSRVD17}.

\begin{equation}
    \label{eq:gate_bilstm}
    \mathcal{H}_{q}={\rm BiLSTM}(E_{q})
\end{equation}
\begin{equation}
    \label{eq:question_gate_w}
    \mathbf{w}_{ij}=\sigma\left(\mathbf{W}_q\left[u_i^l||\mathcal{H}_{q_{j}}\right]\right)
\end{equation}
\begin{equation}
    \label{eq:question_gate_softmax}
    \alpha_{ij}={\rm softmax}(\mathbf{w}_{ij})
\end{equation}
  \begin{equation}
    \label{eq:question_gate_query_sum}
    q_i^l=\sum_{j=1}^{M}\alpha_{ij}\mathcal{H}_{q}
  \end{equation}
  \begin{equation}
    \label{eq:question_gate_query_beta}
    \beta_i^l=\sigma(\mathbf{W}_s[q_i^l||u_i^l])
  \end{equation}
\begin{equation}
    \label{eq:question_gate_query_v}
    \tilde{u}_i^l=\beta_i^l\odot{\rm tanh}(q_i^l)+(1-\beta_i^l)\odot u_i^l
\end{equation}
\begin{equation}
  \label{eq:gate_first}
  w_i^l=\sigma(\textbf{W}_g[\tilde{u}_i^j||g_i^l])
\end{equation}
\begin{equation}
  \label{eq:gate_final}
  u_i^{l+1}=w_i^l\odot{\rm tanh}(\tilde{u}_i^l)+(1-w_i^l)\odot u_i^l
\end{equation}
where $E_q$ is the embedding matrix of the question, $\mathcal{H}_q$ is encoded by an independent Bi-LSTM to keep dimension consistency of node representations, $\sigma$ stands for the ${\rm sigmoid}$ function, and $\odot$ indicates elements-wise multiplication.

\subsubsection{Output Layer}
We represented the candidate drugs with two different types of nodes, \ie $\mathcal{V}_{men}$ and $\mathcal{V}_{can}$, and we calculate the classification scores from them as:

\begin{equation}
  \label{eq:gate_final}
  a^*=f_{can}(\mathcal{U}_{can})+{\rm max}(f_{men}(\mathcal{U}_{men}))
\end{equation}
where $f_{can}$ and $f_{men}$ are two-layer MLPs with {\rm tanh} activation function.
And ${\rm max(\cdot)}$ takes the maximum from $f_{men}(\mathcal{U}_{men})$ in a specific and corresponding to one candidate drug, the sum aggregates then mention node messages across documents and the corresponding candidate node messages from the candidate set.

\section{Experiments}\label{sec:expe}

\subsection{Dataset}

\begin{figure}[t]
    \centering
    \includegraphics[width=\textwidth]{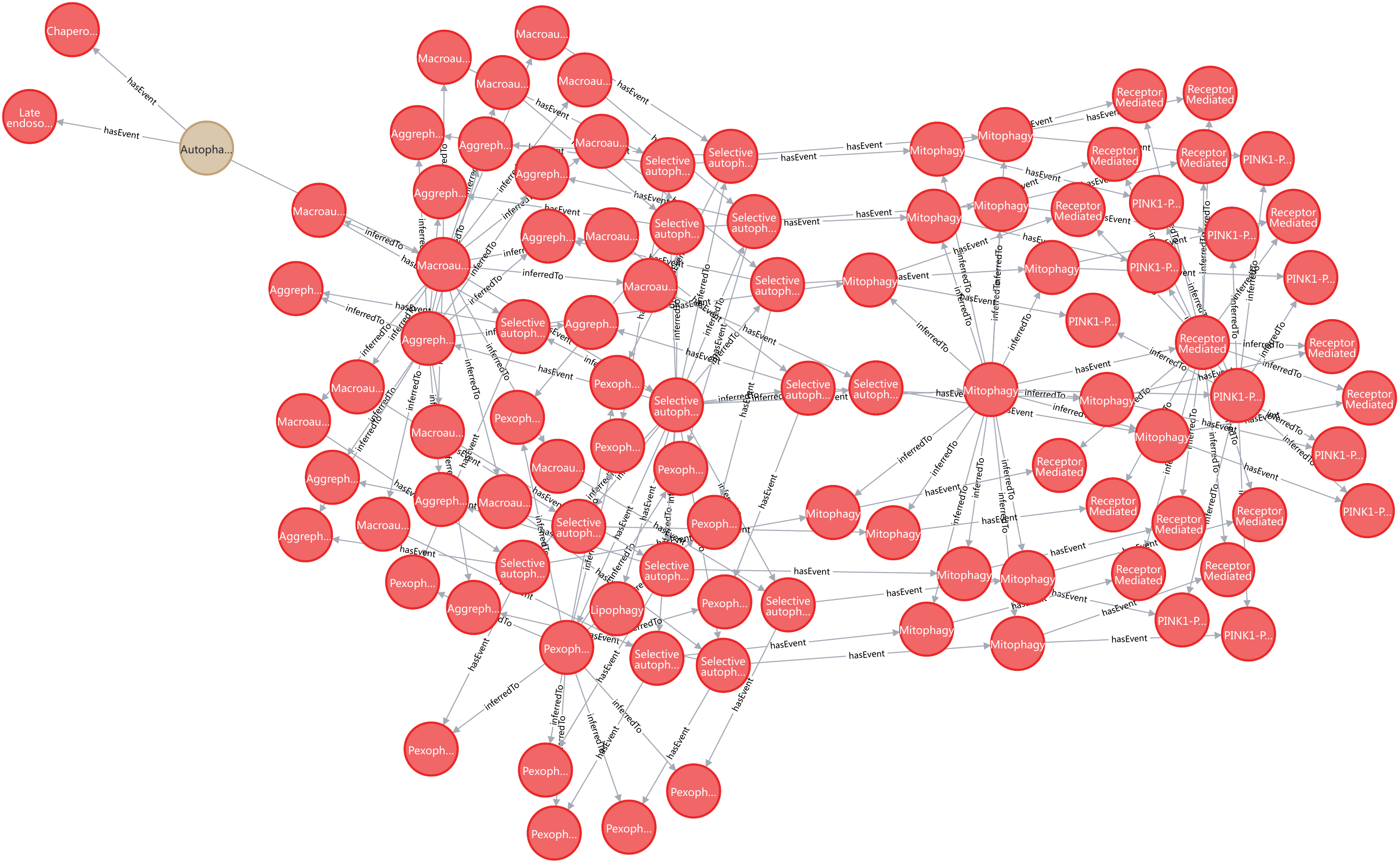}
    \caption{Illustration of biomedical pathways in the \reactome database. The brown color represents a top-level pathway and the others are common pathways. The top-level pathway here is ``Autophagy'', which is an intracellular degradation process by which cells remove and recycle cellular components that are no longer needed or damaged, such as \proteins and organelles. Each common pathway can be further broken down into specific biological reactions. For more information, please refer to the \reactome database.}
    \label{fig:reactiom}
\end{figure}

We extract the knowledge triplets using the \DrugBank \citep{drugbank4}and \reactome \citep{fabregat2018reactome} knowledge bases. \DrugBank database contains relevant knowledge primarily focused on \drugs, and provides 2,689,894 triples $S$ between \drugs, \proteins, and their corresponding knowledge vector embeddings, as shown in Table 2. Meanwhile, \reactome database records biomedical pathways related to protein biomolecules, as depicted in Figure \ref{fig:reactiom}. It provides 20,903 protein pathway information using $(protein_{from},protein_{to})$, which guides the directed connection between the protein reasoning nodes in \emph{MedKGQA}.

\begin{table}[tbp]
\centering
\caption{Sample \drugs-\proteins triplets from the \DrugBank database}
\begin{tabular}{lcc}
\hline
\#\proteins & actions                 & \#\drugs \\ \hline
Q9NY46                     & inhibitor                     & DB00243                 \\
Q99250                     & blocker                       & DB09088                 \\
Q9BYF1                     & inhibitor                     & DB15643                 \\
Q07869                     & activator                     & DB01050                 \\
P0DP23                     & antagonist                    & DB04841                 \\
O60840                     & antagonist                    & DB09236                 \\
P11362                     & inhibitor                     & DB00398                 \\
Q8WXS5                     & agonist                       & DB13746                 \\
O14764                     & positive allosteric modulator & DB13437                 \\
P18507                     & positive allosteric modulator & DB01159                 \\ \hline
\end{tabular}
\label{tab:drugbank}
\end{table}

We used the \qangaroo \MedHop dataset \citep{welbl-etal-2018-constructing} to validate the effectiveness of our model.
Figure~\ref{fig:medhopsample} shows a sample from the original paper.
For this validation, the model was required to predict which drugs can interact with {\slshape Leuprolide}, and candidates and research paper abstracts were presented as reasoning materials.
The sizes of the training, development (dev), and test sets are 1620, 342, and 546, respectively.
To evaluate the model fairly, the test set is not available to the public, which means that the samples in the test set were not available during training. They can be regarded as unobserved or undetected DDI to confirm the capabilities of our model in discovering new DDI.
It is noteworthy that the \emph{MedKGQA} was blindly evaluated by the official committee.

\subsection{Training Settings}

Our proposed model is implemented using PyTorch, trained and evaluated on an instance with $32$ Intel$^\circledR$ Xeon$^\circledR$ Gold-5218 @ 2.3GHz CPU with 256GB RAM, and 2 NVIDIA$^\circledR$ Tesla$^\circledR$ V100 GPU with $64$GB VRAM.
Considering the size of the training set was relatively small, we proposed two different training settings: traditional deep learning training and cross-validation training \citep{zhang2017predicting}.
In the traditional setting, the training and dev sets were independent. We trained the model only on training data and evaluated the model on dev data, preserving the highest-accuracy model within ten training epochs at most.
In the cross-validation process, we combined the training and dev sets into one, then used 9-fold cross-validation.
To avoid overfitting the training in the cross-validation setting, we stopped the training early, in chronological order, and preserved three models for official evaluations.
Each training epoch took approximately 0.5 hours, and evaluating the model took about five minutes.

\subsection{Knowledge Fusion System}
We implemented System 1 based on the open-source framework OpenKE \citep{han2018openke}.
All the knowledge triplets were extracted from \reactome and \DrugBank.
We trained two models \TransE and \TransH with 1000 epochs, and the embedding dimension $h$ was 200.
We saved the entity embeddings for further use, and the missing \emph{drugs} and \emph{proteins} contained in the reasoning graph were replaced with random vectors, as formulated in Equation \ref{eq:coattention_q2sub6}.
The filtered evaluations solely based on System 1 are shown in Table \ref{tab:my-table-trans}, which illustrates two poor performances in hit@1.

\begin{table}[tbp]
    \centering
    \caption{Evaluations of \emph{TransE} and \emph{TransH} in the knowledge fusion system. In order to fully transfer the entity embeddings from the triplets, we do not strictly distinguish the training and test set.}
    \begin{tabular}{cccccc}
    \hline
    \textbf{Trans} & \textbf{MRR} & \textbf{MR} & \textbf{hit@10} & \textbf{hit@3} & \textbf{hit@1} \\ \hline
    TransE         & 0.113952     & 128.004242  & 0.279880        & 0.171930       & 0.003795       \\
    TransH         & 0.144770     & 73.624825   & 0.348708        & 0.228884       & 0.004123       \\ \hline
    \end{tabular}
    \label{tab:my-table-trans}
\end{table}

\subsection{Graph Reasoning System}
In System 2, we tokenized the documents into word sequences with NLTK \citep{DBLP:conf/acl/Bird06} and obtained the subject drug from the question.
We followed the word embedding as \citep{hde}, which contains 300-dimensional GLoVe \citep{pennington-etal-2014-glove} embeddings (with 840B tokens and 2.2M vocabulary size) \citep{pennington-etal-2014-glove} and 100-dimensional character n-gram embeddings \citep{hashimoto-etal-2017-joint}.

The \emph{drugs} and \emph{proteins} were converted into accession numbers in the dataset, such as \textit{DB0001 (Lepirudin)} and \textit{P49122 (Cytotoxin 7)}, which are out-of-vocabulary in the original GLoVe dictionary, so we added all accession numbers into the dictionary and initialized them with random vectors, and set them to be trainable.
The numbers of subject nodes, reasoning nodes, mention nodes, and candidate nodes in the reasoning graph were truncated as 200, 800, 100, and 9, respectively.

\subsection{Hyperparameter Study}

\begin{figure}[H]
    \centering
    \includegraphics[width=0.6\textwidth]{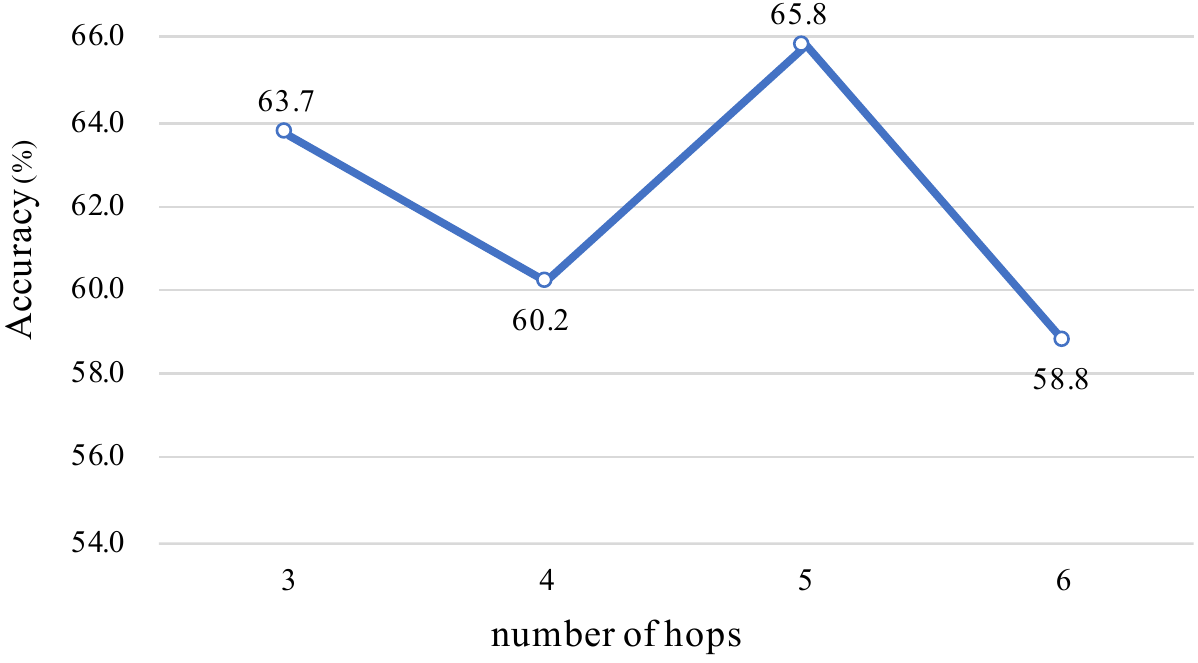}
    \caption{Accuracies in the different number of hops.}
    \label{fig:ana}
\end{figure}

\begin{figure}[H]
    \centering
    \includegraphics[width=0.6\textwidth]{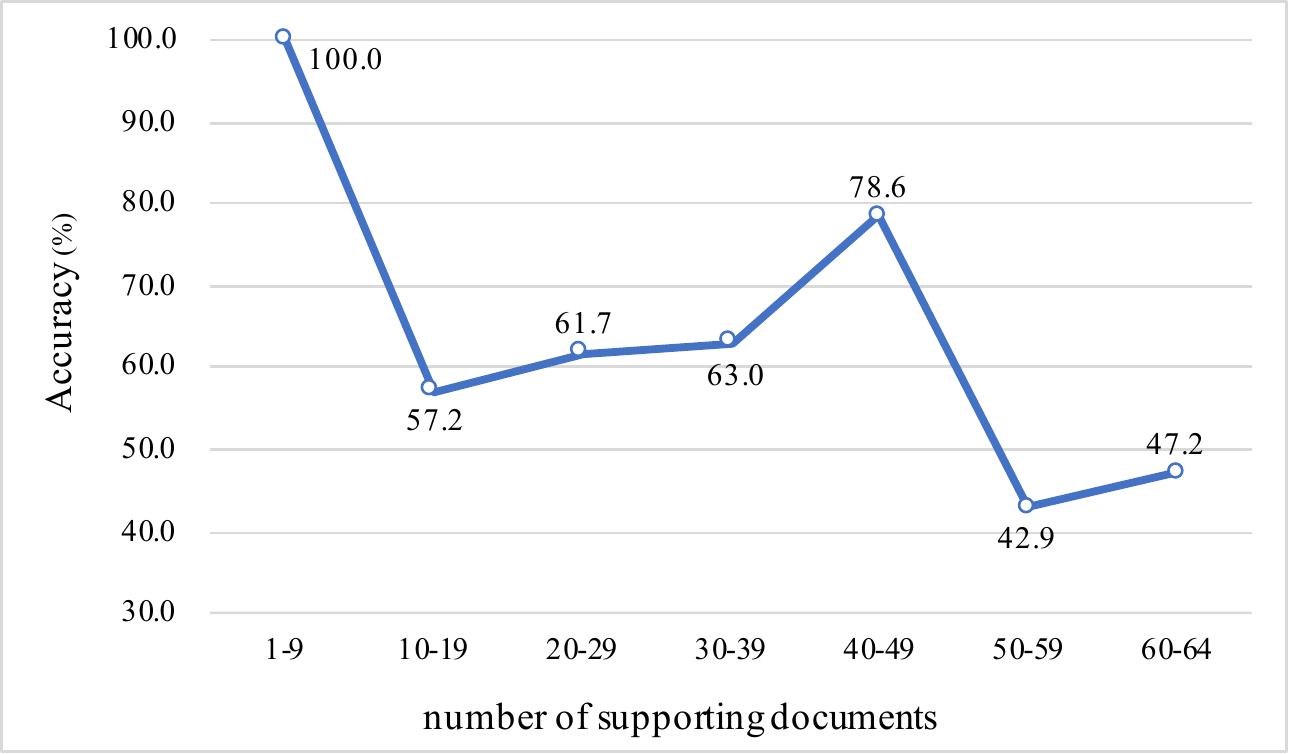}
    \caption{Accuracy in the different number of supporting documents.}
    \label{fig:doc}
\end{figure}

The number of layers (\emph{a.k.a.} hops) in the reasoning graph is a crucial hyperparameter.
Besides the default five layers in \emph{MedKGQA}, we also stacked three, four, and six parameter-sharing layers in System 2.
Apart from the number of layers, the other experimental settings remained consistent, and the results are reported in Figure~\ref{fig:ana}.
\emph{MedKGQA} achieves the highest accuracy in five hops, and the 2nd highest accuracy in three hops.
The accuracy in 4 hops was 60.2\% and 58.8\% in 6 hops.

We also grouped the development set by the number of supporting documents and calculated their accuracies.
Figure~\ref{fig:doc} shows the results in the seven groups.
The highest and second highest accuracy appears in groups 1-9 and 40-49, respectively.
When the number of documents exceeded 50, the accuracy dropped rapidly.

\subsection{Results on \MedHop}

  \newcommand{\tabincell}[2]{\begin{tabular}{@{}#1@{}}#2\end{tabular}}
    \begin{table}[tbp]
      \centering
      \caption{\emph{MedKGQA} performance on \MedHop, and the comparison with the other models. The symbol $*$ stands for the unpublished model, and the symbol -- means that the result of a particular model is unavailable.}
      \begin{tabular}{l|c|c}
      \hline
      \multirow{2}{*}{\textbf{\emph{Single models}}} & \multicolumn{2}{c}{\textbf{Accuracy(\%)}}%
      \\ \cline{2-3}  & \textbf{Dev}             & \textbf{Test}           \\ \hline
      EPAr \citep{DBLP:conf/acl/JiangJCB19}                           & --               & 60.3           \\
      \tabincell{l}{Most Frequent Given Candidate \citep{welbl-etal-2018-constructing}}  & --               & 58.4           \\
      Vanilla CoAttention Model$^*$      & --               & 58.1           \\
      BiDAF \citep{DBLP:conf/iclr/SeoKFH17}                           & --               & 47.8           \\
      ClueReader \citep{cluereader}                 & --               & 46             \\
      \tabincell{l}{Document-cue\citep{welbl-etal-2018-constructing}}                   & --               & 44.9           \\
      \tabincell{l}{FastQA\citep{weissenborn-etal-2017-making}  }                      & --               & 23.1           \\ \hline
      \textbf{MedKGQA}              & \textbf{78.9}   & \textbf{64.8}  \\ \hline
      \end{tabular}
      \label{tab:my-table-result-compare}
   \end{table}

Table \ref{tab:my-table-result-compare} shows the best result of our single model issued by the official evaluation and compares it with the results of the other top seven models on the leaderboard of \MedHop. All models use attention mechanisms to predict DDIs, but only ClueReader~\cite{cluereader} and \emph{MedKGQA} utilize additional knowledge grahs to enhance the MRC capabilities.
We argue that our proposed model, \emph{MedKGQA}, improves the state-of-the-art accuracy on the blind test set from 60.3\% to 64.8\%.

  \begin{table}[htbp]
    \centering
    \caption{The comparison of \emph{MedKGQA} in two training strategies: with or without 9-fold cross-validation. We stopped the training from \#2 to \#4 in chronological order, where the model in \#2 was more competitive.}
    \begin{tabular}{cccc}
    \hline
    \multirow{2}{*}{\#} & \multirow{2}{*}{\textit{\textbf{Cross-validation}}} & \multicolumn{2}{c}{\textbf{Accuracy(\%)}} \\ \cline{3-4}
                       &                                                     & \textbf{Dev}        & \textbf{Test}       \\ \hline
    1                  & No                                                  & 65.8                & 61.2                \\
    2                  & Yes                                                 & \textbf{78.9}       & \textbf{64.8}       \\
    3                  & Yes                                                 & 71.9                & 63.4                \\
    4                  & Yes                                                 & 76.6                & 61.7                \\ \hline
    \end{tabular}
    \label{tab:my-table}
    \end{table}

Table \ref{tab:my-table} shows the performance of our model with/without cross-validation training.
It shows the effectiveness of cross-validation, particularly in \#2, which is our best result among the four trained models.
Even though we used the traditional training setting in \#1, our model's accuracy was still higher than the second-best model by nearly 1\%.

\subsection{Ablation Studies}

We conducted ablation studies based on the \#1 model in Table \ref{tab:my-table}, avoiding samples that appeared in training.
As shown in Table \ref{tab:my-table-abstudy}, the accuracy on the dev set dropped by 15.5\% when the Knowledge Fusion System was deleted.
Completing the nature of drugs was significant during the reasoning course, which supports the claim that our proposed method is simple but effective.
Without the contribution of the Graph Reasoning System and directly taking drug features output from encoders for prediction, the accuracy on the dev set dropped by 18.4\%, which explains that reasoning over graphs can benefit the entity representation learning across documents confirming the possibility of detecting DDI from published scientific literature.
Looking at the above two systems and considering the very poor results solely depending on System 1 as reported in Table \ref{tab:my-table-trans}, we argue that the cooperation of the two systems is crucial to predicting DDI, though the reasoning graph contributes more to the final results.

\begin{table}[htbp]
    \centering
    \caption{The results of the ablation studies on the \MedHop development set. The single module is removed or ineffective in predictions.}
    \begin{tabular}{lcc}
    \hline
    \multirow{2}{*}{\textbf{Model}} & \multicolumn{2}{c}{\textbf{Accuracy(\%)}} \\ \cline{2-3}
        & \textbf{Dev}                    & $\Delta$ \\
        \hline
    Full model             & \multicolumn{1}{c}{\textbf{65.8}} & \textbf{--}       \\
    \quad w/o Knowledge Fusion   & \multicolumn{1}{c}{50.3} & 15.5       \\
    \quad w/o Graph Reasoning    & \multicolumn{1}{c}{47.4} & 18.4       \\
    \quad w/o edge types         & \multicolumn{1}{c}{64.9} & 0.9       \\
    \quad w/o subject nodes      & \multicolumn{1}{c}{64.0} & 1.8       \\
    \quad w/o reasoning nodes    & \multicolumn{1}{c}{62.9} & 2.9       \\
    \quad w/o mention nodes      & \multicolumn{1}{c}{63.2} & 2.6       \\
    \quad w/o candidate nodes    & \multicolumn{1}{c}{63.7} & 2.1       \\ \hline
    \end{tabular}
    \label{tab:my-table-abstudy}
\end{table}

When we did not distinguish the connection types of edges, the prediction accuracy also decreased by 0.9\%, which indicates attention learning in a specific edge can reduce distracting messages passing over a chaotic graph.
Furthermore, the reasoning node made the largest contribution among the four different types of nodes, leading to a 2.9\% reduction in accuracy.
This phenomenon proves that the idea of integrating the PPI chains in DDI prediction is valid.
Subsequently, when we blocked the mention node, candidate node, and subject node, accuracy was reduced by 2.6\%, 2.1\%, and 1.8\%, respectively.
In short, maintaining the integrity of drug pathways in the reasoning graph benefits the comprehensive graph representation learning and enhances DDI predictions.

\subsection{Output Visualizations}
\begin{figure}[htbp]
    \begin{center}
        \subfigure[]
        {\label{fig:scalea}\includegraphics[width=.45\textwidth]{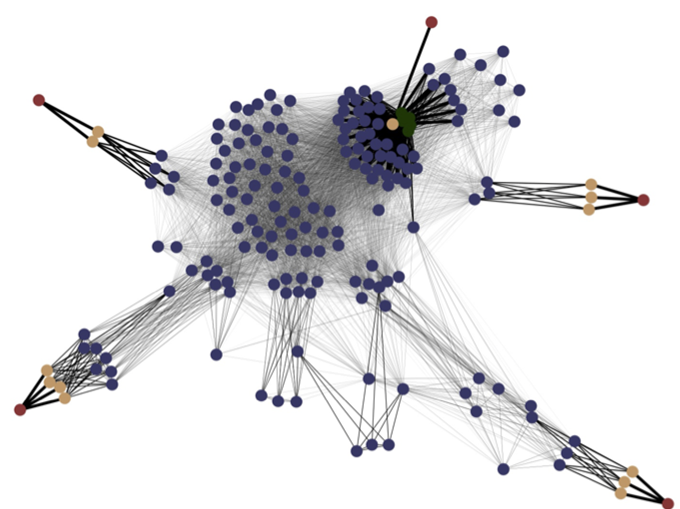}}\hspace{2em}
        \subfigure[]
        {\label{fig:scaleb}\includegraphics[width=.45\textwidth]{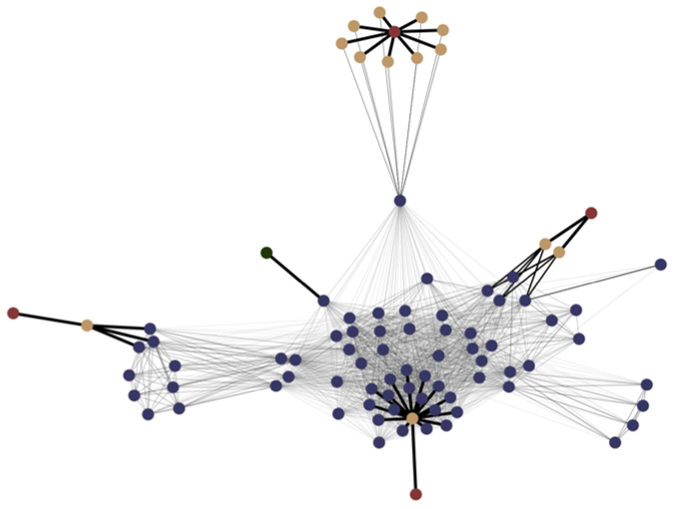}}
        \vfill
        \subfigure[]
        {\label{fig:scalec}\includegraphics[width=.45\textwidth]{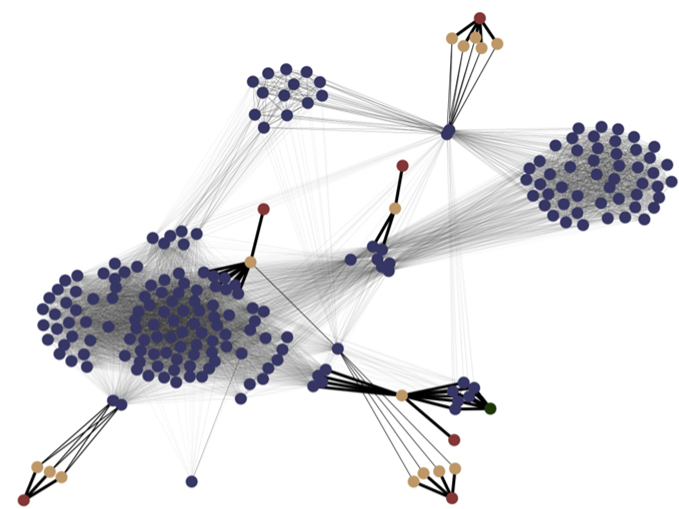}}\hspace{2em}
        \subfigure[]
        {\label{fig:scaled}\includegraphics[width=.45\textwidth]{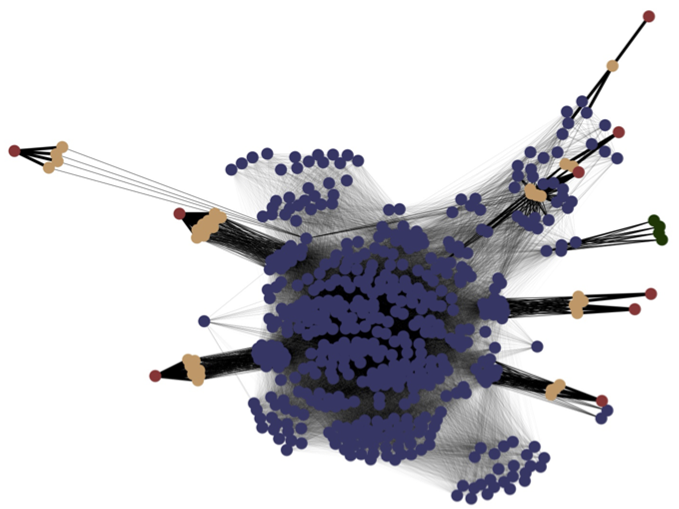}}
        \vfill
        \subfigure[]
        {\label{fig:scalee}\includegraphics[width=.45\textwidth]{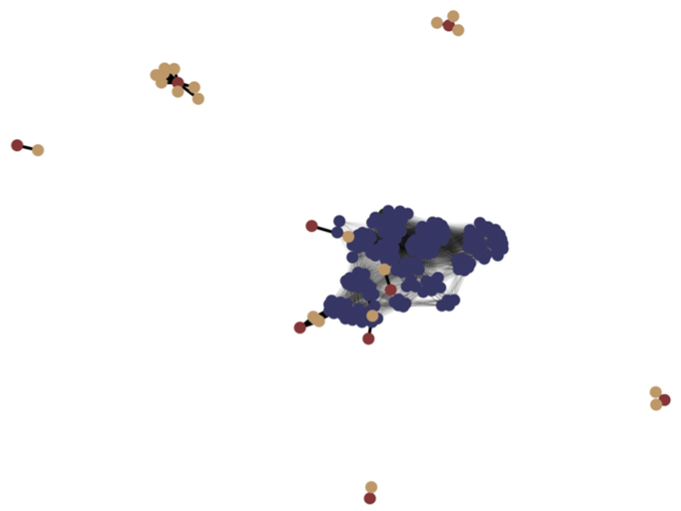}}\hspace{2em}
        \subfigure[]
        {\label{fig:scalef}\includegraphics[width=.45\textwidth]{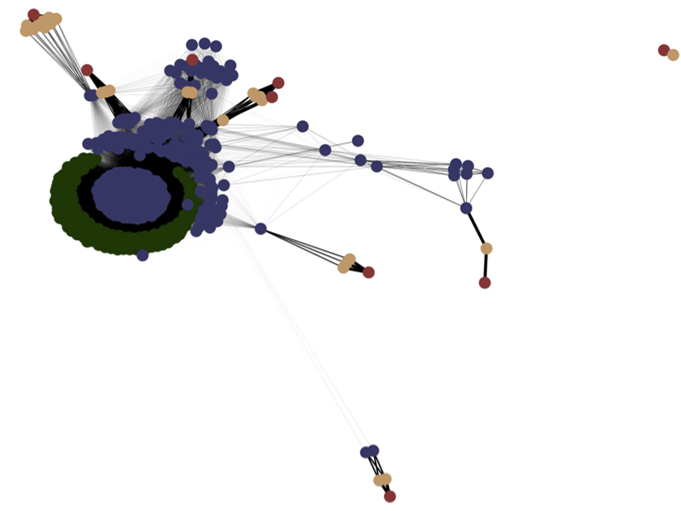}}
    \end{center}
    \caption{Visualization of the reasoning graphs of six randomly selected \MedHop dev set samples. (a)--(d) are the correctly predicted samples, while (e) and (f) are the incorrectly predicted samples.}
    \label{fig:graphrepresentation}
\end{figure}

We visualized the output reasoning graphs from \emph{MedKGQA} with NetworkX\footnote{\url{https://networkx.org}}. Six graphs of randomly selected samples are shown in Figure~\ref{fig:graphrepresentation} where subject nodes are green, reasoning nodes are violet, mention nodes are brown, and candidate nodes are red. The thickness of the edges indicates the importance of the node pair, and the node transparencies in the mention and candidate nodes stand for the selections from the output layer. It can also be seen that in Figures~\ref{fig:scalee} and~\ref{fig:scalef}, the failure to fully extract to the nodes or the failure to establish effective connections led to errors in DDI predictions, thus reducing the prediction accuracy of \emph{MedKGQA}.
Detailed model prediction scores among the mention and candidate drugs can be found at \url{https://medkgqa.github.io/}, and we hope they can contribute to future medical researchers.

\section{Conclusion}\label{sec:con}

In this paper, we propose the \emph{MedKGQA} model, which utilizes a closed-domain multi-document MRC framework to predict DDI with high accuracy. \emph{MedKGQA} establishes four relation-centric heterogeneous nodes for \drugs and \proteins and defines four directed or undirected edges between them based on verified biological pathways. External knowledge is integrated into the vector embeddings to optimize the learning and reasoning of node representations using a multi-relation GAT. Additionally, visualizing correctly predicted samples highlights the necessity of establishing a complete graph, guiding future research towards improved prediction accuracy. Finally, the significant impact of training set size on deep learning-based models is emphasized, and expanding it through data augmentation methods is recommended to improve model prediction accuracy.
We anticipate that our model will make a valuable contribution to the medical research community by combining molecular tasks with NLP techniques. We also aim to extend this methodology to other fields, such as law and physics, which have numerous well-established KBs and document resources.

\section*{Acknowledgments} \label{sec:ack}

The authors would like to thank the UCL machine reading group that created the \qangaroo dataset and their help in evaluating our model.

\bibliography{manuscript}

\begin{thebibliography}{10}
\expandafter\ifx\csname url\endcsname\relax
  \def\url#1{\texttt{#1}}\fi
\expandafter\ifx\csname urlprefix\endcsname\relax\def\urlprefix{URL }\fi
\expandafter\ifx\csname href\endcsname\relax
  \def\href#1#2{#2} \def\path#1{#1}\fi

\bibitem{drugbank4}
V.~Law, C.~Knox, Y.~Djoumbou, T.~Jewison, A.~C. Guo, Y.~Liu, A.~Maciejewski,
  D.~Arndt, M.~Wilson, V.~Neveu, et~al., Drugbank 4.0: shedding new light on
  drug metabolism, Nucleic acids research 42~(D1) (2014) D1091--D1097.

\bibitem{drugbank5}
D.~S. Wishart, Y.~D. Feunang, A.~C. Guo, E.~J. Lo, A.~Marcu, J.~R. Grant,
  T.~Sajed, D.~Johnson, C.~Li, Z.~Sayeeda, et~al., Drugbank 5.0: a major update
  to the drugbank database for 2018, Nucleic acids research 46~(D1) (2018)
  D1074--D1082.

\bibitem{review2-1}
Y.~Qiu, Y.~Zhang, Y.~Deng, S.~Liu, W.~Zhang, A comprehensive review of
  computational methods for drug-drug interaction detection, IEEE/ACM
  transactions on computational biology and bioinformatics 19~(4) (2021)
  1968--1985.

\bibitem{review3-1}
G.~Yenduri, G.~Srivastava, P.~K.~R. Maddikunta, R.~H. Jhaveri, W.~Wang, A.~V.
  Vasilakos, T.~R. Gadekallu, et~al., Generative pre-trained transformer: A
  comprehensive review on enabling technologies, potential applications,
  emerging challenges, and future directions, arXiv preprint arXiv:2305.10435.

\bibitem{review3-2}
R.~Chengoden, N.~Victor, T.~Huynh-The, G.~Yenduri, R.~H. Jhaveri, M.~Alazab,
  S.~Bhattacharya, P.~Hegde, P.~K.~R. Maddikunta, T.~R. Gadekallu, Metaverse
  for healthcare: A survey on potential applications, challenges and future
  directions, IEEE Access.

\bibitem{Cohen2004}
K.~B. Cohen, L.~Hunter, Natural language processing and systems biology,
  Artificial intelligence methods and tools for systems biology (2004)
  147--173.

\bibitem{rohani2019drug}
N.~Rohani, C.~Eslahchi, Drug-drug interaction predicting by neural network
  using integrated similarity, Scientific Reports 9~(1) (2019) 1--11.

\bibitem{lazarou1998incidence}
J.~Lazarou, B.~H. Pomeranz, P.~N. Corey, Incidence of adverse drug reactions in
  hospitalized patients: a meta-analysis of prospective studies, Jama 279~(15)
  (1998) 1200--1205.

\bibitem{review2-2}
S.~Liu, Y.~Zhang, Y.~Cui, Y.~Qiu, Y.~Deng, Z.~Zhang, W.~Zhang, Enhancing
  drug-drug interaction prediction using deep attention neural networks,
  IEEE/ACM transactions on computational biology and Bioinformatics 20~(2)
  (2022) 976--985.

\bibitem{review2-4}
Y.~Deng, X.~Xu, Y.~Qiu, J.~Xia, W.~Zhang, S.~Liu, A multimodal deep learning
  framework for predicting drug--drug interaction events, Bioinformatics
  36~(15) (2020) 4316--4322.

\bibitem{hanton2007preclinical}
G.~Hanton, Preclinical cardiac safety assessment of drugs, Drugs in R \& D 8
  (2007) 213--228.

\bibitem{welbl-etal-2018-constructing}
J.~Welbl, P.~Stenetorp, S.~Riedel, Constructing datasets for multi-hop reading
  comprehension across documents, Transactions of the Association for
  Computational Linguistics 6 (2018) 287--302.

\bibitem{doi:10.1086/601720}
D.~R. Swanson, Undiscovered public knowledge, The Library Quarterly 56~(2)
  (1986) 103--118.

\bibitem{10.1007/978-3-030-60450-9_3}
X.~Li, Y.~Liu, S.~Ju, Z.~Xie, Dynamic reasoning network for multi-hop question
  answering, in: Natural Language Processing and Chinese Computing: 9th CCF
  International Conference, NLPCC 2020, Zhengzhou, China, October 14--18, 2020,
  Proceedings, Part I 9, Springer, 2020, pp. 29--40.

\bibitem{zhang2009predicting}
L.~Zhang, Y.~Zhang, P.~Zhao, S.-M. Huang, Predicting drug--drug interactions:
  an fda perspective, The AAPS journal 11 (2009) 300--306.

\bibitem{beida}
A.~Yang, Q.~Wang, J.~Liu, K.~Liu, Y.~Lyu, H.~Wu, Q.~She, S.~Li, Enhancing
  pre-trained language representations with rich knowledge for machine reading
  comprehension, in: Proceedings of the 57th Annual Meeting of the Association
  for Computational Linguistics, 2019, pp. 2346--2357.

\bibitem{DBLP:conf/acl/InkpenZLCW18}
Q.~Chen, X.~Zhu, Z.-H. Ling, D.~Inkpen, S.~Wei, Neural natural language
  inference models enhanced with external knowledge, in: Proceedings of the
  56th Annual Meeting of the Association for Computational Linguistics (Volume
  1: Long Papers), 2018, pp. 2406--2417.

\bibitem{DBLP:conf/semeval/WangSZSL18}
L.~Wang, M.~Sun, W.~Zhao, K.~Shen, J.~Liu, Yuanfudao at semeval-2018 task 11:
  Three-way attention and relational knowledge for commonsense machine
  comprehension, in: Proceedings of The 12th International Workshop on Semantic
  Evaluation, 2018, pp. 758--762.

\bibitem{cho2014learning}
K.~Cho, B.~van Merri{\"e}nboer, {\c{C}}.~Gul{\c{c}}ehre, D.~Bahdanau,
  F.~Bougares, H.~Schwenk, Y.~Bengio, Learning phrase representations using rnn
  encoder--decoder for statistical machine translation, in: Proceedings of the
  2014 Conference on Empirical Methods in Natural Language Processing (EMNLP),
  2014, pp. 1724--1734.

\bibitem{wang2018multi}
Y.~Wang, K.~Liu, J.~Liu, W.~He, Y.~Lyu, H.~Wu, S.~Li, H.~Wang, Multi-passage
  machine reading comprehension with cross-passage answer verification, in:
  Proceedings of the 56th Annual Meeting of the Association for Computational
  Linguistics (Volume 1: Long Papers), 2018, pp. 1918--1927.

\bibitem{hochreiter1997long}
S.~Hochreiter, J.~Schmidhuber, Long short-term memory, Neural computation 9~(8)
  (1997) 1735--1780.

\bibitem{zhou2016attention}
P.~Zhou, W.~Shi, J.~Tian, Z.~Qi, B.~Li, H.~Hao, B.~Xu, Attention-based
  bidirectional long short-term memory networks for relation classification,
  in: Proceedings of the 54th annual meeting of the association for
  computational linguistics (volume 2: Short papers), 2016, pp. 207--212.

\bibitem{clark2018simple}
C.~Clark, M.~Gardner, Simple and effective multi-paragraph reading
  comprehension, in: Proceedings of the 56th Annual Meeting of the Association
  for Computational Linguistics (Volume 1: Long Papers), 2018, pp. 845--855.

\bibitem{brown2020language}
T.~Brown, B.~Mann, N.~Ryder, M.~Subbiah, J.~D. Kaplan, P.~Dhariwal,
  A.~Neelakantan, P.~Shyam, G.~Sastry, A.~Askell, et~al., Language models are
  few-shot learners, Advances in neural information processing systems 33
  (2020) 1877--1901.

\bibitem{lewis2020bart}
M.~Lewis, Y.~Liu, N.~Goyal, M.~Ghazvininejad, A.~Mohamed, O.~Levy, V.~Stoyanov,
  L.~Zettlemoyer, Bart: Denoising sequence-to-sequence pre-training for natural
  language generation, translation, and comprehension, in: Proceedings of the
  58th Annual Meeting of the Association for Computational Linguistics, 2020,
  pp. 7871--7880.

\bibitem{transformer}
A.~Vaswani, N.~Shazeer, N.~Parmar, J.~Uszkoreit, L.~Jones, A.~N. Gomez,
  {\L}.~Kaiser, I.~Polosukhin, Attention is all you need, Advances in neural
  information processing systems 30.

\bibitem{beltagy2020longformer}
I.~Beltagy, M.~E. Peters, A.~Cohan, Longformer: The long-document transformer,
  arXiv preprint arXiv:2004.05150.

\bibitem{zaheer2020big}
M.~Zaheer, G.~Guruganesh, K.~A. Dubey, J.~Ainslie, C.~Alberti, S.~Ontanon,
  P.~Pham, A.~Ravula, Q.~Wang, L.~Yang, et~al., Big bird: Transformers for
  longer sequences, Advances in neural information processing systems 33 (2020)
  17283--17297.

\bibitem{de-cao-etal-2019-question}
N.~De~Cao, W.~Aziz, I.~Titov, Question answering by reasoning across documents
  with graph convolutional networks, in: Proceedings of the 2019 Conference of
  the North American Chapter of the Association for Computational Linguistics:
  Human Language Technologies, Volume 1 (Long and Short Papers), 2019, pp.
  2306--2317.

\bibitem{cao-etal-2019-bag}
Y.~Cao, M.~Fang, D.~Tao, Bag: Bi-directional attention entity graph
  convolutional network for multi-hop reasoning question answering, in:
  Proceedings of the 2019 Conference of the North American Chapter of the
  Association for Computational Linguistics: Human Language Technologies,
  Volume 1 (Long and Short Papers), 2019, pp. 357--362.

\bibitem{cluereader}
P.~Gao, F.~Gao, P.~Wang, J.-C. Ni, F.~Wang, H.~Fujita, Cluereader:
  Heterogeneous graph attention network for multi-hop machine reading
  comprehension, Electronics 12~(14) (2023) 3183.

\bibitem{review2-5}
W.~Zhang, K.~Jing, F.~Huang, Y.~Chen, B.~Li, J.~Li, J.~Gong, Sflln: a sparse
  feature learning ensemble method with linear neighborhood regularization for
  predicting drug--drug interactions, Information Sciences 497 (2019) 189--201.

\bibitem{zhang2015label}
P.~Zhang, F.~Wang, J.~Hu, R.~Sorrentino, Label propagation prediction of
  drug-drug interactions based on clinical side effects, Scientific reports
  5~(1) (2015) 12339.

\bibitem{zhang2017predicting}
W.~Zhang, Y.~Chen, F.~Liu, F.~Luo, G.~Tian, X.~Li, Predicting potential
  drug-drug interactions by integrating chemical, biological, phenotypic and
  network data, BMC bioinformatics 18 (2017) 1--12.

\bibitem{rohani2020iscmf}
N.~Rohani, C.~Eslahchi, A.~Katanforoush, Iscmf: integrated
  similarity-constrained matrix factorization for drug--drug interaction
  prediction, Network Modeling Analysis in Health Informatics and
  Bioinformatics 9 (2020) 1--8.

\bibitem{DBLP:conf/acl/JiangJCB19}
Y.~Jiang, N.~Joshi, Y.-C. Chen, M.~Bansal, Explore, propose, and assemble: An
  interpretable model for multi-hop reading comprehension, in: Proceedings of
  the 57th Annual Meeting of the Association for Computational Linguistics,
  2019, pp. 2714--2725.

\bibitem{review2-3}
H.~Fu, F.~Huang, X.~Liu, Y.~Qiu, W.~Zhang, Mvgcn: data integration through
  multi-view graph convolutional network for predicting links in biomedical
  bipartite networks, Bioinformatics 38~(2) (2022) 426--434.

\bibitem{transe}
A.~Bordes, N.~Usunier, A.~Garcia-Duran, J.~Weston, O.~Yakhnenko, Translating
  embeddings for modeling multi-relational data, Vol.~26, 2013.

\bibitem{transh}
Z.~Wang, J.~Zhang, J.~Feng, Z.~Chen, Knowledge graph embedding by translating
  on hyperplanes, in: Proceedings of the AAAI conference on artificial
  intelligence, Vol.~28, 2014.

\bibitem{lstm}
S.~Hochreiter, J.~Schmidhuber, Long short-term memory, Neural computation 9~(8)
  (1997) 1735--1780.

\bibitem{fabregat2018reactome}
A.~Fabregat, S.~Jupe, L.~Matthews, K.~Sidiropoulos, M.~Gillespie, P.~Garapati,
  R.~Haw, B.~Jassal, F.~Korninger, B.~May, et~al., The reactome pathway
  knowledgebase, Nucleic acids research 46~(D1) (2018) D649--D655.

\bibitem{hde}
M.~Tu, G.~Wang, J.~Huang, Y.~Tang, X.~He, B.~Zhou, Multi-hop reading
  comprehension across multiple documents by reasoning over heterogeneous
  graphs, in: Proceedings of the 57th Annual Meeting of the Association for
  Computational Linguistics, 2019, pp. 2704--2713.

\bibitem{DBLP:conf/iclr/ZhongXKS19}
V.~Zhong, C.~Xiong, N.~S. Keskar, R.~Socher, Coarse-grain fine-grain
  coattention network for multi-evidence question answering, in: International
  Conference on Learning Representations, 2019.

\bibitem{gat}
P.~Veli{\v{c}}kovi{\'c}, G.~Cucurull, A.~Casanova, A.~Romero, P.~Li{\`o},
  Y.~Bengio, Graph attention networks, in: International Conference on Learning
  Representations, 2018.

\bibitem{DBLP:conf/ijcai/TangSMXYL20}
Z.~Tang, Y.~Shen, X.~Ma, W.~Xu, J.~Yu, W.~Lu, Multi-hop reading comprehension
  across documents with path-based graph convolutional network, in: Proceedings
  of the Twenty-Ninth International Conference on International Joint
  Conferences on Artificial Intelligence, 2021, pp. 3905--3911.

\bibitem{DBLP:conf/icml/GilmerSRVD17}
J.~Gilmer, S.~S. Schoenholz, P.~F. Riley, O.~Vinyals, G.~E. Dahl, Neural
  message passing for quantum chemistry, in: International conference on
  machine learning, PMLR, 2017, pp. 1263--1272.

\bibitem{han2018openke}
X.~Han, S.~Cao, X.~Lv, Y.~Lin, Z.~Liu, M.~Sun, J.~Li, Openke: An open toolkit
  for knowledge embedding, in: Proceedings of the 2018 conference on empirical
  methods in natural language processing: system demonstrations, 2018, pp.
  139--144.

\bibitem{DBLP:conf/acl/Bird06}
E.~Loper, S.~Bird, Nltk: The natural language toolkit, in: Proceedings of the
  ACL-02 Workshop on Effective Tools and Methodologies for Teaching Natural
  Language Processing and Computational Linguistics, 2002, pp. 63--70.

\bibitem{pennington-etal-2014-glove}
J.~Pennington, R.~Socher, C.~D. Manning, Glove: Global vectors for word
  representation, in: Proceedings of the 2014 conference on empirical methods
  in natural language processing (EMNLP), 2014, pp. 1532--1543.

\bibitem{hashimoto-etal-2017-joint}
K.~Hashimoto, C.~Xiong, Y.~Tsuruoka, R.~Socher, A joint many-task model:
  Growing a neural network for multiple nlp tasks, in: Proceedings of the 2017
  Conference on Empirical Methods in Natural Language Processing, 2017, pp.
  1923--1933.

\bibitem{DBLP:conf/iclr/SeoKFH17}
M.~Seo, A.~Kembhavi, A.~Farhadi, H.~Hajishirzi, Bidirectional attention flow
  for machine comprehension, in: International Conference on Learning
  Representations, 2017.

\bibitem{weissenborn-etal-2017-making}
D.~Weissenborn, G.~Wiese, L.~Seiffe, Making neural qa as simple as possible but
  not simpler, in: Proceedings of the 21st Conference on Computational Natural
  Language Learning (CoNLL 2017), 2017, pp. 271--280.

\end{thebibliography}

\end{document}